%% file: main.tex
\documentclass{article}
\usepackage{spconf,amsmath,graphicx}
\usepackage{enumitem}
\setlist{nosep, leftmargin=14pt}
\usepackage{xcolor}

\newcommand\blfootnote[1]{
  \begingroup
  \renewcommand\thefootnote{}\footnote{#1}
  \addtocounter{footnote}{-1}
  \endgroup
}

\title{MultiStar: Instance Segmentation of Overlapping Objects With Star-convex Polygons}
\name{Florin C. Walter, Sebastian Damrich, Fred A. Hamprecht}
\address{HCI/IWR, University of Heidelberg, Germany}

\begin{document}

\maketitle

\begin{abstract}
Instance segmentation of overlapping objects in biomedical images remains a largely unsolved problem. We take up this challenge and present \textit{MultiStar}, an extension to the popular instance segmentation method \textit{StarDist}. The key novelty of our method is that we identify pixels at which objects overlap and use this information to improve proposal sampling and to avoid suppressing proposals of truly overlapping objects. This allows us to apply the ideas of \textit{StarDist} to images with overlapping objects, while incurring only a small overhead compared to the established method. \textit{MultiStar} shows promising results on two datasets and has the advantage of using a simple and easy to train network architecture. 
\end{abstract}


\blfootnote{© 2021 IEEE.  Personal use of this material is permitted.  Permission from IEEE must be obtained for all other uses, in any current or future media, including reprinting/republishing this material for advertising or promotional purposes, creating new collective works, for resale or redistribution to servers or lists, or reuse of any copyrighted component of this work in other works.}
\section{Introduction}
\label{sec:intro}
Instance segmentation is the image analysis task of identifying distinct objects of the same category and assigning unique instance labels to the associated pixels. While difficult in itself, this problem becomes even more challenging when the objects are clustered and appear to overlap in the image projection. Many applications require segmentation of the full overlapping objects, which means that multiple instance labels need to be assigned to pixels in the object overlap. \\ \indent 
A common biomedical application is the segmentation of cells in microscopy images, where objects can be dense and seemingly overlap with each other. Reliable automated processing of such images reduces the need for human experts.  \\ \indent
In recent years, segmentation of overlapping objects has been addressed by a number of methods. \cite{isoodl} and \cite{isoodl_v2} lift the label space to 3D and shear the segmentation masks,  yielding a non-overlapping representation. \cite{diskmask} introduces an end-to-end approach with an encoder-decoder network, and \cite{irnet} uses instance relation interaction. \cite{molnar} presents a variational method for segmentation of near-circular shaped objects. \\ \indent
Instance segmentation of dense but non-overlapping objects has been demonstrated in a very effective and elegant way by \textit{StarDist} \cite{stardist}\cite{stardist3d}, which parameterizes objects by star-convex polygons. With \textit{MultiStar} we extend this method to images with overlapping objects, by additionally predicting object overlap. In doing so, we can identify pixels at which a star-convex parameterization is unambiguously possible and account for the predicted overlap in the non-maximum suppression. The main strength of \textit{MultiStar} is that while being a very simple and straight-forward modification of the successful \textit{StarDist}, it greatly extends its application area.
\section{Method}
\label{sec:methods}
\begin{figure}[t]
  \centering
  \def\svgwidth{8.5cm}
  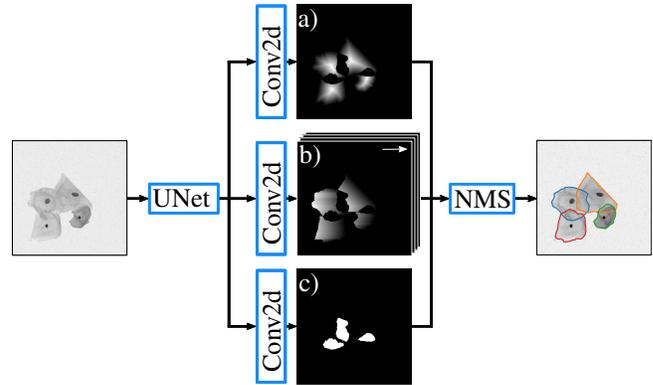
\caption{The model predicts \textbf{a)} \textit{Object Probability}, \textbf{b)} \textit{Star Distances} (one of 32 directions shown) and \textbf{c)} \textit{Overlap Probability}. Overlap areas are masked in a) and b). Incorporating the predicted overlap in the non-maximum suppression (NMS) allows to generate overlapping segmentations.}
\label{fig:model}
\end{figure}
\subsection{Review of \textit{StarDist}}
\label{ssec:review}
\textit{StarDist} \cite{stardist} is a proposal based method. For every pixel $p$ it predicts a polygon to capture the pixel's object instance. The polygons are parameterized by the \textit{Star Distances}, defined as the Euclidean distances from $p$ to the polygon vertices, along a fixed number of equiangular radial directions. Proposals are generated at positions that are sampled according to the predicted \textit{Object Probability},  defined inside objects as the normalized Euclidean distance transform and in the background as zero. The \textit{Object Probability} furthermore acts as confidence score in the subsequent non-maximum suppression (NMS), which determines the final set of proposals representing the instance segmentation. \\ \indent
NMS requires setting a threshold on the intersection over union (IoU) between two proposals, above which the less confident proposal is suppressed. This threshold has to be low enough to avoid multiple detections of the same object. But a low threshold also prevents the final proposals from overlapping, even if the ground truth objects do overlap. \textit{StarDist} therefore needs to trade off avoiding to detect false positives against detecting truly overlapping objects. This makes it difficult to determine the optimal IoU threshold in the NMS. \\ \indent
Another important point when considering images with object overlap is that \textit{Star Distances} and \textit{Object Probability} are ill-defined at overlap pixels, because it is not clear to which object they should refer. \\ \indent
\subsection{\textit{MultiStar}}
\label{ssec:multistar}
\begin{figure}[t]
  \centering
  \def\svgwidth{8.5cm}
  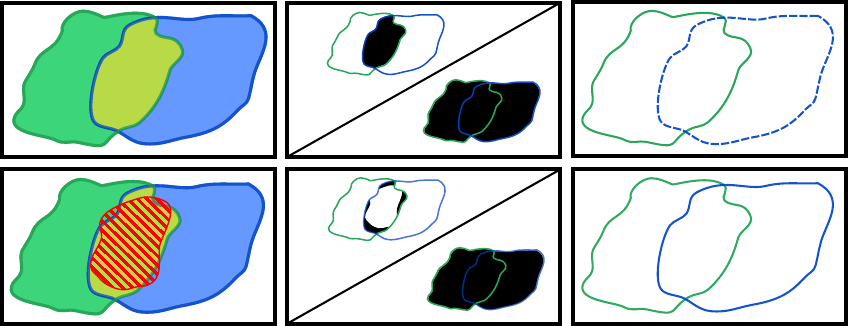
\caption{\textbf{a)} \textit{StarDist} does not predict the object overlap (yellow), which leads to \textbf{b)} large IoU and thus \textbf{c)} suppression of the less confident proposal (dotted contour). \textbf{d)} \textit{Multistar} predicts the overlap (red, hashed), \textbf{e)} excludes it from the intersection and due to the low IoU \textbf{f)} both proposals are accepted.}
\label{fig:iou}
\end{figure}

We solve both of these issues by additionally predicting the \textit{Overlap Probability} $ P_{over} $ (Fig. \ref{fig:model} c)). Its ground truth value is $ 1 $ at pixels where at least two objects overlap and $ 0 $ elsewhere. The prediction can take intermediate values, reflecting different degrees of certainty. \\ \indent
Given the predicted \textit{Object Probability} $ P_{obj} $, we sample a proposal at pixel $ p $ with the thresholded probability
\begin{equation}
    P_{proposal}(p) \propto P_{obj}(p) \cdot (1 - P_{over}(p))
\end{equation}
as opposed to simply sampling from $ P_{obj} (p)$. In the ideal case of correct overlap probabilities $ P_{over}(p) \in \{0, 1\} $, this corresponds to sampling only at non-overlap pixels, where the proposal parameterization is well-defined. For intermediate values, the probability to sample at $p$ is reduced according to how confidently the model predicts overlap at $p$. \\ \indent
In the NMS, \textit{MultiStar} avoids suppressing detections of truly overlapping objects by excluding the predicted overlap from the intersection $I$ of two proposals $A$ and $B$ 
\begin{equation}
    I \equiv \sum_{p \in A \cap B} (1 - P_{over}(p))
\label{eq:intersection}
\end{equation}
and computing the IoU with the usual definition of the union. The rationale here is that when two proposals overlap in an area where the model predicts object overlap (high $P_{over}$), the two proposals are probably correct detections of overlapping objects and not multiple detections of the same object. If computed with (\ref{eq:intersection}), the IoU in this case is low and neither of the two objects is suppressed (Fig. \ref{fig:iou}). Conversely, if two proposals overlap just slightly where the model does not predict any overlap, this indicates false positives. Hence, it is possible to choose a low IoU threshold without suppressing detections of truly overlapping objects. \\ \indent
\textit{MultiStar} can detect an unlimited number of objects involved in the overlap. It can however not detect objects that are fully overlapped by other objects, because it does not allow to sample proposals at predicted overlap pixels.
\subsection{Model}
\label{ssec:model}
We use a generic UNet \cite{unet} with $3$ output branches (Fig. \ref{fig:model}), very similar to \cite{stardist} but with one additional output branch for the \textit{Overlap Probability} prediction. \\ \indent
The UNet backbone has $5$ levels with $16$, $32$, $64$, $128$ and $ 256 $ channels. Each down-/upsampling block consists of two $3\times3$ convolutions with Batch Normalization \cite{batchnorm}, ReLU activations and subsequent $2\times2$ max-pooling or upsampling. The UNet output has $256$ channels and ReLU activations. \\ \indent
We append three output branches for the three prediction features, consisting of a single convolutional layer each. In the \textit{Object Probability} and \textit{Overlap Probability} branches we use a single output channel with sigmoid activations. In the \textit{Star Distances} branch we use $32$ output channels for the $32$ radial directions and ReLU activations.
\subsection{Training}
\label{ssec:training}
Inspired by \cite{multitaskloss}, we optimize over the network parameters $ \theta $ and the task uncertainties $ \sigma_i $ to minimize the regularized weighted sum of the three network output losses
\begin{equation}
\begin{aligned}
    L(\theta, \sigma_i) &= \frac{1}{\sigma_{over}^{2}} L_{over}(\theta) + \frac{1}{\sigma_{obj}^{2}} L_{obj}(\theta) \\& + \frac{1}{\sigma_{dist}^{2}} L_{dist}(\theta) + \log(\sigma_{over} \sigma_{obj} \sigma_{dist}).
\end{aligned}
\end{equation}
$ L_{over} $ and $ L_{obj} $ are binary cross-entropy losses. $L_{dist} $ is the mean absolute difference between the predicted and true \textit{Star Distances} with every pixel's contribution weighted by its true \textit{Object Probability}. Pixels at which ground truth objects overlap are excluded from $ L_{obj} $ and $ L_{dist} $. The model parameters and task uncertainties are jointly optimized with Adam \cite{adam} and learning rate $ 10^{-4} $. We apply random flips, rotations and elastic deformations to augment the training data.

\section{Experiments}
\label{sec:experiments}
\subsection{Evaluation metrics}
\label{ssec:metrics}
 We use the metrics described in \cite{isbi14_1}: For matched predictions and ground truth objects with dice coefficient greater than $0.7$, we determine the average dice coefficient DC (higher is better) and pixel-based true positive and false positive rates TP\textsubscript{p} and FP\textsubscript{p}. The object-based false negative rate FN\textsubscript{o} accounts for ground truth objects without matched prediction with dice coefficient greater than $0.7$. \\ \indent
 The above metrics do not penalize false positive detections. This is problematic, as an excessive amount thereof limits the usefulness of a segmentation result. Hence, we also compute the average precision $AP = \frac{TP}{TP + FP + FN} $ per image and average over all images. A pair of a prediction and a ground truth object is considered a true positive TP if its IoU is above a threshold $\tau$, otherwise as a false positive FP.
\subsection{Datasets and Results}
\label{ssec:datasetsresults}
\begin{figure}[t]
  \centering
  \def\svgwidth{8.5cm}
  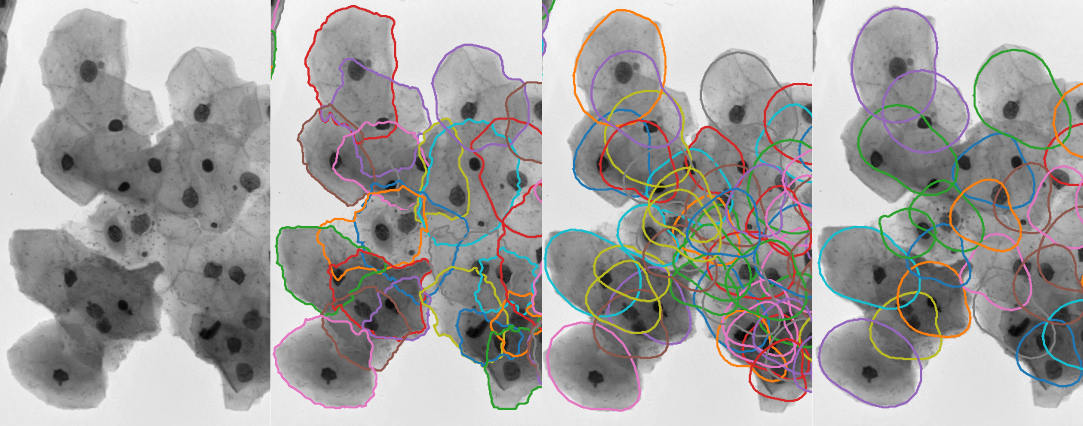
\caption{Example image from the \textbf{OSC-ISBI} dataset. From left to right: original image, ground truth segmentation, \textit{MultiStar1} segmentation, \textit{MultiStar2} segmentation.}
\label{fig:isbi15}
\end{figure}

\begin{figure}[h]
  \centering
  \def\svgwidth{8.5cm}
  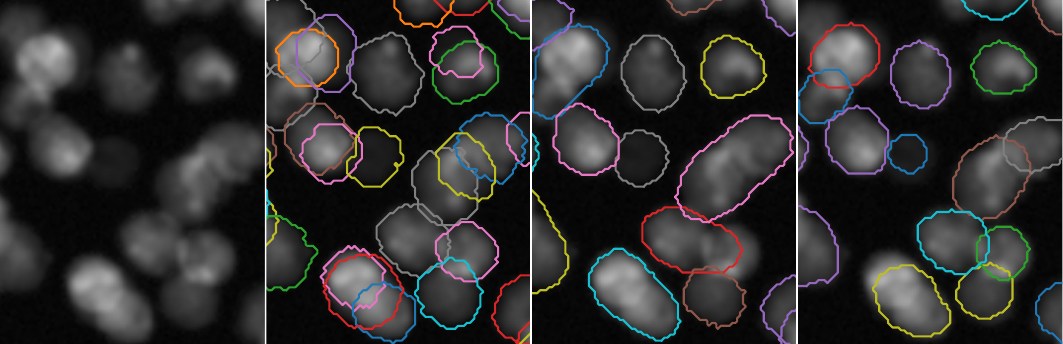
\caption{Example image from the \textbf{DSB-OV} dataset. From left to right: original image, ground truth segmentation, pretrained \textit{StarDist} segmentation, \textit{MultiStar} segmentation.}
\label{fig:dsb18}
\end{figure}

The \textbf{OSC-ISBI} dataset contains images from the first and second \textit{Overlapping Cervical Cytology Image Segmentation Challenge} \cite{isbi14_1}\cite{isbi14_2}. As in [1-3], we train on the $945$ synthetic images from \cite{isbi14_1} and the $8$ training images from \cite{isbi14_2} and evaluate on the $9$ test images from \cite{isbi14_2}. \\ \indent
\begin{table}[b]
\centering
\scalebox{0.7}{
\begin{tabular}{|c|c|c|c|c|c|} 
\hline
& $ \mathbf{DC} \uparrow $ & $ \mathbf{FN_o} \downarrow $ & $ \mathbf{TP_p} \uparrow $ & $ \mathbf{FP_p} \downarrow $ & $ \mathbf{AP} \uparrow $\\ [0.5ex]
\hline\hline
\textit{ISOO\textsubscript{DL}} \cite{isoodl} & $.86 \pm .07 $ & $ .37 \pm .14 $ & $ .90 \pm .11 $ & $ .001 \pm .001 $ & - \\ \hline
\textit{ISOO\textsubscript{DL}\textsuperscript{V2}} \cite{isoodl_v2} & $ .90 \pm .08 $ & $ .29 \pm .15 $ & $ .90 \pm .11 $ & $ .001 \pm .001 $ & - \\ \hline
\textit{Diskmask} \cite{diskmask} & $ .90 \pm .08 $ & $ .22 \pm .13 $ & $ .90 \pm .11 $ & $ .001 \pm .001 $ & - \\ \hline
\textit{\textbf{MultiStar1}} & $ .86 \pm .07 $ & $ .31 \pm .13 $ & $ .83 \pm .10 $ & $ .001 \pm .001 $ & $ .21 \pm .06 $ \\ \hline
\textit{\textbf{MultiStar2}} & $ .85 \pm .07 $ & $ .42 \pm .16 $ & $ .82 \pm .10 $ & $ .001 \pm .001 $ & $ .47 \pm .13 $ \\
\hline
\end{tabular}
}
\caption{Results on the test images of the \textbf{OSC-ISBI} dataset.}
\label{table:isbi15}
\end{table}
Our experiments show that the optimal choice of the threshold $ \rho $ on $ P_{proposal} $ and the IoU threshold $ \nu $ in the NMS depends on which metrics are evaluated. We obtain the best results with respect to DC, TP\textsubscript{p}, FP\textsubscript{p} and FN\textsubscript{o} with $ \rho = 0.2 $ and $ \nu = 0.5 $ (\textit{MultiStar1} in Table \ref{table:isbi15}). Despite its good scores, this setting generates many false positive detections, see Fig. \ref{fig:isbi15}. By considering only AP with $ \tau = 0.5 $, we obtain the best results with $ \rho = 0.1 $ and $ \nu = 0.2 $ (\textit{MultiStar2} in Table \ref{table:isbi15}). Only FN\textsubscript{o} is significantly worse in this setting than in \textit{MultiStar1}. Nevertheless, qualitatively the segmentation results look more useful for a practitioner, see Fig. 3. This suggests that optimizing DC, TP\textsubscript{p}, FP\textsubscript{p} and FN\textsubscript{o} alone might not be sufficient to identify a useful segmentation method for overlapping instances. \\ \indent
Note that \textit{MultiStar} is trained using only the cytoplasm annotations, whereas \cite{isoodl}\cite{isoodl_v2}\cite{diskmask} also use nuclei annotations. Remarkably, we still manage to achieve competitive DC, FN\textsubscript{o} and TP\textsubscript{p} scores, while only falling behind on the TP\textsubscript{p}. \\ \indent
Our second dataset is based on the real microscopy images of cell nuclei from various organisms that were used in \cite{stardist} to establish the performance of \textit{StarDist}\footnote{github.com/mpicbg-csbd/stardist/releases/download/0.1.0/dsb2018.zip subset of the \textit{Kaggle Data Science Bowl 2018} dataset available at kaggle.com/c/data-science-bowl-2018}. Since these images do not contain object overlap, we create the synthetic dataset \textbf{DSB-OV}\footnote{Code for our model and the DSB-OV dataset is publicly available at https://github.com/overlapping-instances/MultiStar.} by randomly replicating, flipping, rotating and shifting objects in the images and adding up the intensities, such that at least $ 15 \% $ of the object pixels in every image are in the overlap of multiple objects. \\ \indent
Based on AP with different thresholds $\tau$, we compare \textit{MultiStar} with the pretrained \textit{StarDist} model from \cite{stardist} on \textbf{DSB-OV} in Table \ref{table:dsb18}. We find the optimal values $ \rho = 0.3 $ and $ \nu = 0.1 $ for \textit{MultiStar}, and $ \rho = 0.5 $ and $ \nu = 0.5 $ for \textit{StarDist}. For all but the strictest $\tau$ \textit{MultiStar} significantly outperforms \textit{StarDist}. As argued above, the optimal $ \nu $ is significantly higher for \textit{StarDist} than for \textit{MultiStar}, since \textit{StarDist} does not explicitly model overlapping instances. Even at this high $ \nu $, \textit{StarDist} tends to merge overlapping objects, while \textit{MultiStar} separates them much better, see Fig. \ref{fig:dsb18}. For \textit{MultiStar} a low $ \nu $ is enough to retain overlapping proposals while avoiding false positives.
\begin{table}[h!]
\centering
\scalebox{0.7}{
\begin{tabular}{|c|c|c|c|c|c|} 
\hline
$ \tau $ & $ 0.4 $ & $ 0.5 $ & $ 0.6 $ & $ 0.7 $ & $ 0.8 $ \\ [0.5ex]
\hline\hline
\textit{\textbf{StarDist}} & $ .62 \pm .15 $ & $ .55 \pm .16 $ & $ .46 \pm .16 $ & $ .34 \pm .14 $ & $ .19 \pm .10 $ \\ \hline
\textit{\textbf{MultiStar}} & $ .73 \pm .13 $ & $ .66 \pm .16 $ & $ .54 \pm .21 $ & $ .33 \pm .22 $ & $ .15 \pm .18 $ \\
\hline
\end{tabular}
}
\caption{AP results of the pretrained \textit{StarDist} and \textit{MultiStar} with optimal $\rho$ and $\nu$ on the test images of the \textbf{DSB-OV} dataset.}
\label{table:dsb18}
\end{table}

\section{Discussion}
We introduced a very simple but effective extension to the popular method \textit{StarDist} \cite{stardist}, which extends its applicability to images with strongly overlapping objects. With a much simpler architecture and without exploiting cell nuclei annotation for training, we are only slightly below the current state of the art on the \textbf{OSC-ISBI} dataset. Compared to the original \textit{StarDist}, we achieve a substantial boost in performance on images of overlapping cell nuclei, which we believe could be further improved by using a more elaborate training procedure.

\section{Compliance with Ethical Standards}
\label{sec:ethics}
This research study was conducted retrospectively using human and animal subject data made available in open access by \cite{isbi14_1}, \cite{isbi14_2} and \cite{caicedo2019nucleus}.
\section{Acknowledgments}
\label{sec:acknowledgments}
Funded by the \textit{Deutsche Forschungsgemeinschaft} (DFG, German Research Foundation) – Projektnummer 240245660 - SFB 1129.
\bibliographystyle{IEEEbib}
\bibliography{refs}

\end{document}

%% file: model.pdf_tex
%% Creator: Inkscape inkscape 0.92.5, www.inkscape.org
%% Accompanies image file 'model.pdf' (pdf, eps, ps)
\begingroup%
  \makeatletter%
  \providecommand\color[2][]{%
    \errmessage{(Inkscape) Color is used for the text in Inkscape, but the package 'color.sty' is not loaded}%
    \renewcommand\color[2][]{}%
  }%
  \providecommand\transparent[1]{%
    \errmessage{(Inkscape) Transparency is used (non-zero) for the text in Inkscape, but the package 'transparent.sty' is not loaded}%
    \renewcommand\transparent[1]{}%
  }%
  \providecommand\rotatebox[2]{#2}%
  \newcommand*\fsize{\dimexpr\f@size pt\relax}%
  \newcommand*\lineheight[1]{\fontsize{\fsize}{#1\fsize}\selectfont}%
  \ifx\svgwidth\undefined%
    \setlength{\unitlength}{789.0919302bp}%
    \ifx\svgscale\undefined%
      \relax%
    \else%
      \setlength{\unitlength}{\unitlength * \real{\svgscale}}%
    \fi%
  \else%
    \setlength{\unitlength}{\svgwidth}%
  \fi%
  \global\let\svgwidth\undefined%
  \global\let\svgscale\undefined%
  \makeatother%
  \begin{picture}(1,0.59378336)%
    \lineheight{1}%
    \setlength\tabcolsep{0pt}%
    \put(0,0){\includegraphics[width=\unitlength,page=1]{model.pdf}}%
    \put(0.29958802,0.40799085){\color[rgb]{0,0,0}\rotatebox{90.7127527}{\makebox(0,0)[lt]{\begin{minipage}{0.13873892\unitlength}\raggedright \end{minipage}}}}%
    \put(0.05914493,0.47951914){\color[rgb]{0,0,0}\rotatebox{90.41312867}{\makebox(0,0)[lt]{\begin{minipage}{0.15965433\unitlength}\raggedright \end{minipage}}}}%
    \put(0,0){\includegraphics[width=\unitlength,page=1]{model.pdf}}%
    \put(0.44687104,0.55890105){\color[rgb]{1,1,1}\makebox(0,0)[lt]{\lineheight{1.25}\smash{\begin{tabular}[t]{l}a)\end{tabular}}}}%
    \put(0.44904565,0.34517979){\color[rgb]{1,1,1}\makebox(0,0)[lt]{\lineheight{1.25}\smash{\begin{tabular}[t]{l}b)\end{tabular}}}}%
    \put(0.44822612,0.1447324){\color[rgb]{1,1,1}\makebox(0,0)[lt]{\lineheight{1.25}\smash{\begin{tabular}[t]{l}c)\end{tabular}}}}%
    \put(0.41986829,0.01810511){\color[rgb]{0,0,0}\rotatebox{90}{\makebox(0,0)[lt]{\lineheight{1.25}\smash{\begin{tabular}[t]{l}Conv2d\end{tabular}}}}}%
    \put(0.41922513,0.43227403){\color[rgb]{0,0,0}\rotatebox{90}{\makebox(0,0)[lt]{\lineheight{1.25}\smash{\begin{tabular}[t]{l}Conv2d\end{tabular}}}}}%
    \put(0.41986829,0.21822115){\color[rgb]{0,0,0}\rotatebox{90}{\makebox(0,0)[lt]{\lineheight{1.25}\smash{\begin{tabular}[t]{l}Conv2d\end{tabular}}}}}%
    \put(0.21990954,0.27633471){\color[rgb]{0,0,0}\makebox(0,0)[lt]{\lineheight{1.25}\smash{\begin{tabular}[t]{l}UNet\end{tabular}}}}%
    \put(0.69174693,0.2760841){\color[rgb]{0,0,0}\makebox(0,0)[lt]{\lineheight{1.25}\smash{\begin{tabular}[t]{l}NMS\end{tabular}}}}%
  \end{picture}%
\endgroup%

%% file: iou.pdf_tex
%% Creator: Inkscape inkscape 0.92.5, www.inkscape.org
%% Accompanies image file 'iou.pdf' (pdf, eps, ps)
\begingroup%
  \makeatletter%
  \providecommand\color[2][]{%
    \errmessage{(Inkscape) Color is used for the text in Inkscape, but the package 'color.sty' is not loaded}%
    \renewcommand\color[2][]{}%
  }%
  \providecommand\transparent[1]{%
    \errmessage{(Inkscape) Transparency is used (non-zero) for the text in Inkscape, but the package 'transparent.sty' is not loaded}%
    \renewcommand\transparent[1]{}%
  }%
  \providecommand\rotatebox[2]{#2}%
  \newcommand*\fsize{\dimexpr\f@size pt\relax}%
  \newcommand*\lineheight[1]{\fontsize{\fsize}{#1\fsize}\selectfont}%
  \ifx\svgwidth\undefined%
    \setlength{\unitlength}{407.04371368bp}%
    \ifx\svgscale\undefined%
      \relax%
    \else%
      \setlength{\unitlength}{\unitlength * \real{\svgscale}}%
    \fi%
  \else%
    \setlength{\unitlength}{\svgwidth}%
  \fi%
  \global\let\svgwidth\undefined%
  \global\let\svgscale\undefined%
  \makeatother%
  \begin{picture}(1,0.38396825)%
    \lineheight{1}%
    \setlength\tabcolsep{0pt}%
    \put(0,0){\includegraphics[width=\unitlength,page=1]{iou.pdf}}%
    \put(0.00661515,0.34185795){\color[rgb]{0,0,0}\makebox(0,0)[lt]{\lineheight{1.25}\smash{\begin{tabular}[t]{l}a)\end{tabular}}}}%
    \put(0.34472961,0.34197787){\color[rgb]{0,0,0}\makebox(0,0)[lt]{\lineheight{1.25}\smash{\begin{tabular}[t]{l}b)\end{tabular}}}}%
    \put(0.67931694,0.34379656){\color[rgb]{0,0,0}\makebox(0,0)[lt]{\lineheight{1.25}\smash{\begin{tabular}[t]{l}c)\end{tabular}}}}%
    \put(0.00571546,0.1452741){\color[rgb]{0,0,0}\makebox(0,0)[lt]{\lineheight{1.25}\smash{\begin{tabular}[t]{l}d)\end{tabular}}}}%
    \put(0.34257712,0.14658404){\color[rgb]{0,0,0}\makebox(0,0)[lt]{\lineheight{1.25}\smash{\begin{tabular}[t]{l}e)\end{tabular}}}}%
    \put(0.6814798,0.14639009){\color[rgb]{0,0,0}\makebox(0,0)[lt]{\lineheight{1.25}\smash{\begin{tabular}[t]{l}f)\end{tabular}}}}%
  \end{picture}%
\endgroup%

%% file: isbi15_multistar.pdf_tex
%% Creator: Inkscape inkscape 0.92.5, www.inkscape.org
%% Accompanies image file 'isbi15_multistar.pdf' (pdf, eps, ps)
\begingroup%
  \makeatletter%
  \providecommand\color[2][]{%
    \errmessage{(Inkscape) Color is used for the text in Inkscape, but the package 'color.sty' is not loaded}%
    \renewcommand\color[2][]{}%
  }%
  \providecommand\transparent[1]{%
    \errmessage{(Inkscape) Transparency is used (non-zero) for the text in Inkscape, but the package 'transparent.sty' is not loaded}%
    \renewcommand\transparent[1]{}%
  }%
  \providecommand\rotatebox[2]{#2}%
  \newcommand*\fsize{\dimexpr\f@size pt\relax}%
  \newcommand*\lineheight[1]{\fontsize{\fsize}{#1\fsize}\selectfont}%
  \ifx\svgwidth\undefined%
    \setlength{\unitlength}{519.7545187bp}%
    \ifx\svgscale\undefined%
      \relax%
    \else%
      \setlength{\unitlength}{\unitlength * \real{\svgscale}}%
    \fi%
  \else%
    \setlength{\unitlength}{\svgwidth}%
  \fi%
  \global\let\svgwidth\undefined%
  \global\let\svgscale\undefined%
  \makeatother%
  \begin{picture}(1,0.39295137)%
    \lineheight{1}%
    \setlength\tabcolsep{0pt}%
    \put(0,0){\includegraphics[width=\unitlength,page=1]{isbi15_multistar.pdf}}%
  \end{picture}%
\endgroup%

%% file: dsb18_multistar.pdf_tex
%% Creator: Inkscape inkscape 0.92.5, www.inkscape.org
%% Accompanies image file 'dsb18_multistar.pdf' (pdf, eps, ps)
\begingroup%
  \makeatletter%
  \providecommand\color[2][]{%
    \errmessage{(Inkscape) Color is used for the text in Inkscape, but the package 'color.sty' is not loaded}%
    \renewcommand\color[2][]{}%
  }%
  \providecommand\transparent[1]{%
    \errmessage{(Inkscape) Transparency is used (non-zero) for the text in Inkscape, but the package 'transparent.sty' is not loaded}%
    \renewcommand\transparent[1]{}%
  }%
  \providecommand\rotatebox[2]{#2}%
  \newcommand*\fsize{\dimexpr\f@size pt\relax}%
  \newcommand*\lineheight[1]{\fontsize{\fsize}{#1\fsize}\selectfont}%
  \ifx\svgwidth\undefined%
    \setlength{\unitlength}{509.9976757bp}%
    \ifx\svgscale\undefined%
      \relax%
    \else%
      \setlength{\unitlength}{\unitlength * \real{\svgscale}}%
    \fi%
  \else%
    \setlength{\unitlength}{\svgwidth}%
  \fi%
  \global\let\svgwidth\undefined%
  \global\let\svgscale\undefined%
  \makeatother%
  \begin{picture}(1,0.32355796)%
    \lineheight{1}%
    \setlength\tabcolsep{0pt}%
    \put(0,0){\includegraphics[width=\unitlength,page=1]{dsb18_multistar.pdf}}%
  \end{picture}%
\endgroup%